\documentclass{article}
\usepackage{spconf,amsmath,graphicx,hyperref}
\usepackage{algorithm}
\usepackage{algpseudocode} 
\usepackage{booktabs}
\usepackage{adjustbox}
\usepackage{hyperref}
\usepackage{amsmath}
\usepackage{amssymb} 

\title{StyleDecoupler: Generalizable Artistic Style Disentanglement }
\name{Zexi Jia, Jinchao Zhang*, Jie Zhou}
\address{ Wechat AI, Tencent, China}

\begin{document}
%
\maketitle

\begin{abstract} 
Representing artistic style is challenging due to its deep entanglement with semantic content. We propose StyleDecoupler, an information-theoretic framework that leverages a key insight: multi-modal vision models encode both style and content, while uni-modal models suppress style to focus on content-invariant features. By using uni-modal representations as content-only references, we isolate pure style features from multi-modal embeddings through mutual information minimization. StyleDecoupler operates as a plug-and-play module on frozen Vision-Language Models without fine-tuning. We also introduce WeART, a large-scale benchmark of 280K artworks across 152 styles and 1,556 artists. Experiments show state-of-the-art performance on style retrieval across WeART and WikiART, while enabling applications like style relationship mapping and generative model evaluation. We release our method and dataset at \href{https://huggingface.co/datasets/wechat-prcap/weart}{this url}.

\end{abstract}


\vspace{-3pt}
\section{Introduction}

Representing artistic style, the unique synthesis of form, color, and composition that distinguishes a Van Gogh from a Monet, is a fundamental goal of computer vision. Progress, however, is impeded by a central challenge: style is deeply entangled with semantic content within modern neural representations. This entanglement causes current models~\cite{wright2022artfid, somepalli2024measuring, jia2025visual, huang2025semantic, jia2025imitation} to be brittle; they perform well on styles seen during training but fail to generalize to the vast and subtle diversity of global art, struggling to distinguish a subject from the manner of its depiction.

We uncover a critical insight: multi-modal and uni-modal vision models encode fundamentally different aspects of images. Multi-modal models like CLIP~\cite{radford2021learning, zhai2023sigmoid}, trained on image-text pairs, learn rich representations that naturally capture both content and style. In contrast, uni-modal models like DINOv2~\cite{oquab2023dinov2}, trained with aggressive augmentations to achieve view invariance, actively suppress stylistic variations to focus on content-invariant features. This divergence, typically seen as a limitation, becomes our key to disentanglement.

Our core contribution, StyleDecoupler, exploits this complementary nature: we use uni-modal representations as a content-only reference to isolate pure style features from multi-modal embeddings. Through an information-theoretic framework, we project out content-correlated dimensions while preserving style-specific information, effectively purifying the style signal. This approach is both principled and practical. StyleDecoupler operates as a lightweight, plug-and-play module on frozen VLM features, eliminating the need for costly retraining.

To rigorously evaluate style disentanglement and catalyze future research, we construct WeART, a new large-scale dataset of over 280,000 artworks. It addresses critical gaps in existing benchmarks by offering balanced coverage of underrepresented categories, high-quality annotations verified by experts, and a hierarchical structure that supports cross-cultural analysis.

Armed with our disentangled representations, we demonstrate significant downstream impact. Our method enables fine-grained style retrieval, uncovers meaningful latent manifolds of artistic movements, and provides reliable metrics for evaluating generative models' stylistic fidelity. In summary, our contributions are: (1) a novel insight into the complementary nature of multi-modal and uni-modal representations for style-content disentanglement; (2) the lightweight StyleDecoupler framework that leverages this insight; and (3) the large-scale WeART benchmark. Together, they set a new state-of-the-art in style representation and open new avenues for computational art understanding.

\begin{figure*}[h!]
\begin{center}
\includegraphics[width=0.8\linewidth]{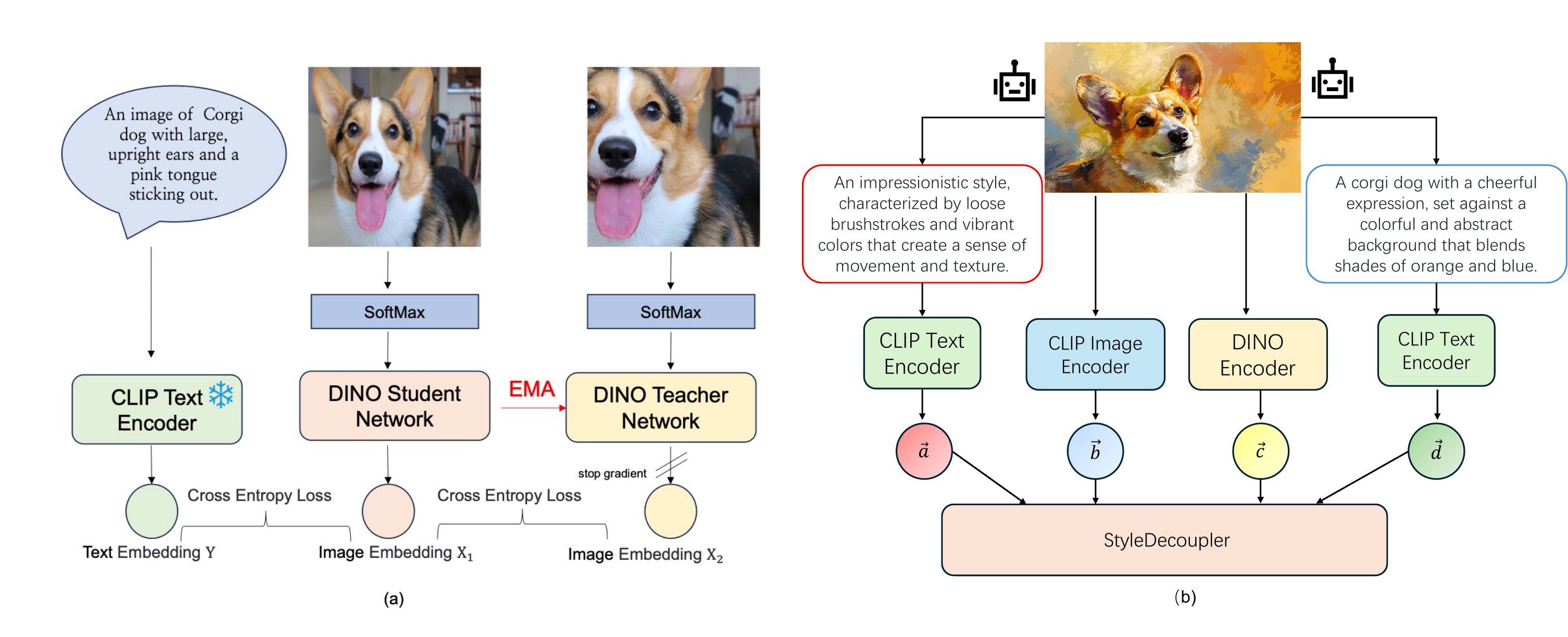}
\end{center}
\caption{Overview of our information-theoretic style disentanglement framework. 
(a) \textbf{Feature Space Alignment:} We align DINO features with the CLIP embedding space using knowledge distillation. 
(b) \textbf{Style Disentanglement:} Guided by GPT-4o generated descriptions, we extract and separate style vectors from content vectors.}
\label{fig:method}
\label{fig:method}
\end{figure*}

\vspace{-5pt}

\section{Related Work}

\noindent\textbf{Style Representation Learning.} Early efforts in style representation relied on hand-crafted statistical features~\cite{graham2012statistics, lun2015elements}. More recent deep learning methods, such as ArtFID~\cite{wright2022artfid} and CSD~\cite{somepalli2024measuring}, have shown success using classification and contrastive learning, respectively. A common thread in these approaches is their reliance on domain-specific training or fine-tuning on art datasets. This inherently limits their ability to generalize to unseen artistic styles~\cite{bin2024gallerygpt, khan2024ai}. and makes them dependent on the scope and biases of their training data. In stark contrast, our approach requires no artistic fine-tuning, making it inherently more robust and scalable.

\noindent\textbf{Style Disentanglement.} Disentangling representations~\cite{jia2025control,huang2025artfrd,huang2025mcid} within VLMs like CLIP~\cite{radford2021learning} is an active area of research. However, existing methods are often designed for specific downstream tasks. For instance, approaches like StyleDiffusion~\cite{wang2023stylediffusion} and Hi-CMD~\cite{choi2020hi} focus on disentanglement for generative modeling ~\cite{shah2024ziplora, ouyang2025k} and typically require architectural modifications or complex, task-specific training regimes. Our work introduces a different paradigm: a model-agnostic disentanglement module. StyleDecoupler offers a universal, plug-and-play solution applicable to pre-trained VLMs.

\vspace{-5pt}
\section{Methodology}

\subsection{Style-Content Disentanglement Framework}

We formalize style representation through information theory to disentangle style from content in multimodal systems. Given image modality $X_i$ and text modality $X_t$, any visual representation decomposes into style $S$ (artistic techniques, visual patterns) and content $C$ (objects, scenes, semantic meaning). Under this partition, style-related mutual information can be isolated:
\begin{equation}
I(X_i, X_t; S) = I(X_i; X_t) - I(X_i, X_t; C)
\label{eq:style_content_relation}
\end{equation}

While Vision-Language Models maximize cross-modal mutual information $I(Z_i ; Z_t)$, the Data Processing Inequality reveals that fine-tuning irreversibly loses style information: $I(X_i, X_t ; S) \geq I(Z_i, Z_t ; S) \geq I(\hat{Z}_i, \hat{Z}_t ; S)$. Our approach instead preserves the original representation space while explicitly removing content components through orthogonal projection.

CLIP and DINO encode complementary information due to distinct training objectives. CLIP preserves both content and style describable in natural language ($I(Z_i; C) + I(Z_i; S_d)$), while DINO's augmentation-based training preserves primarily content ($I(Z_i; C)$). This enables our decoupling strategy:
\begin{equation}
I(Z_i^{CLIP}; S) \approx I(Z_i^{CLIP}; S, C) - I(Z_i^{DINO}; C)
\label{eq:style_extraction}
\end{equation}

\begin{figure*}[h]
\begin{center}
\includegraphics[width=0.8\linewidth]{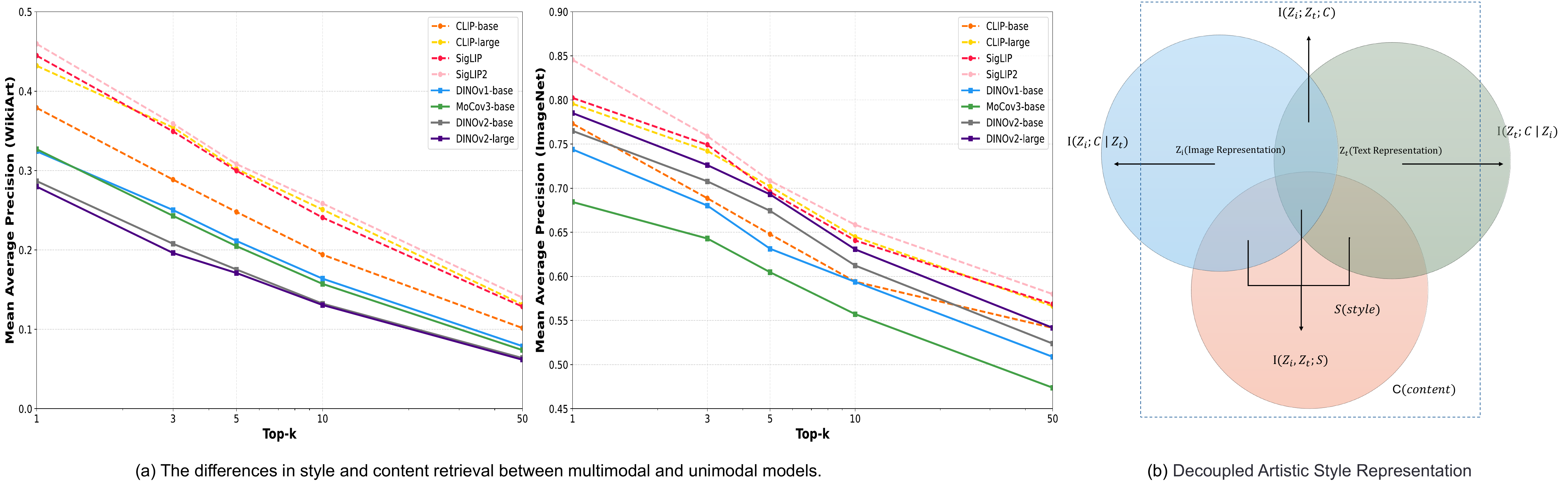}
\end{center}
\caption{Our method is motivated by the observation that VLMs, unlike unimodal models, perform robustly on both natural and artistic retrieval. We leverage this by introducing an information-theoretic framework that explicitly disentangles the VLM's native style and content features.}
\label{fig:mov}
\label{fig:mov}
\end{figure*}

\begin{figure*}[h]
\begin{center}
\includegraphics[width=0.8\linewidth]{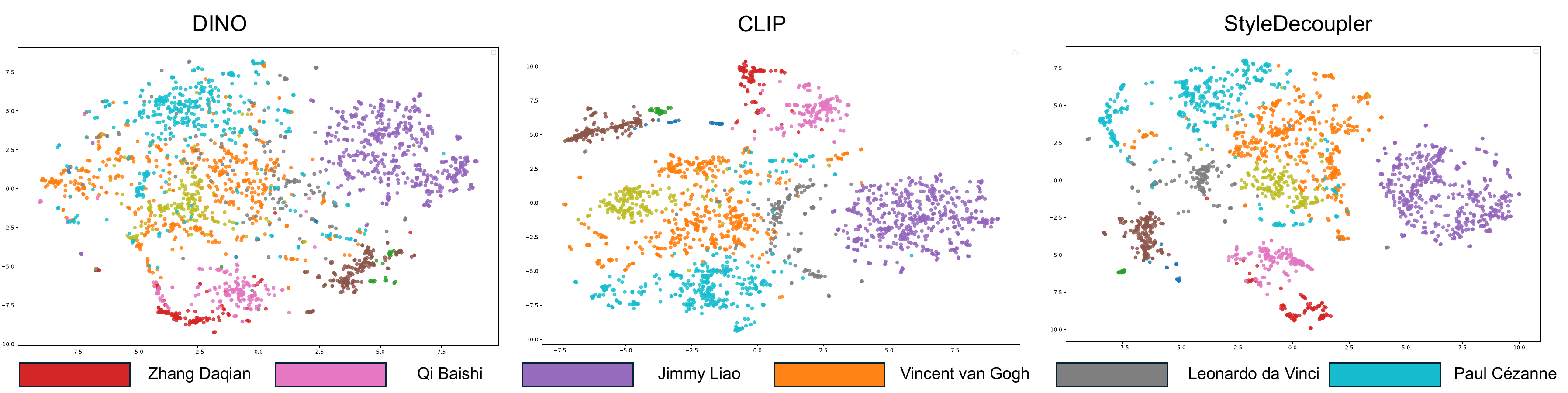}
\end{center}
   \caption{The t-SNE visualization of representation models.
}
\label{fig:tsne}
\end{figure*}

To implement this, we first align DINOv2 with CLIP's feature space using 500,000 image-text pairs from CC3M~\cite{changpinyo2021conceptual}. The alignment objective combines self-distillation with cross-modal constraint:
\begin{equation}
\mathcal{L} = \mathcal{H}(P_{teacher}(i), P_{student}(i)) + \mathcal{H}(CLIP_{text}(t), P_{student}(i))
\end{equation}

As shown in Figure~\ref{fig:method}, for each image, we extract four features: CLIP image features $\bar{b}$ (style+content), aligned DINO features $\bar{c}$ (content), and CLIP text features from GPT-4o generated style ($\bar{a}$) and content ($\bar{d}$) descriptions. The combined representations are:
\begin{equation}
\vec{s_r} = \text{Norm}(\bar{a} + \bar{b}), \quad \vec{c_r} = \text{Norm}(\bar{d} + \bar{c})
\end{equation}

Pure style representation is obtained via confidence-weighted orthogonal projection:
\begin{equation}
\vec{s}_{pure} = \text{Norm}\left(\vec{s_r} - \alpha \cdot \frac{\vec{s_r} \cdot \vec{c_r}}{|\vec{c_r}|^2} \cdot \vec{c_r}\right)
\end{equation}
where $\alpha = \max(0, 1 - \text{sim}(\vec{s_r}, \vec{c_r}))$ allows partial preservation of intrinsic style-content correlations. Figure~\ref{fig:mov}(a) validates that cross-modal models capture richer style information than single-modal approaches, confirming our theoretical foundation.

\subsection{The WeART Benchmark}

To address the limitations of existing art datasets, such as the Western-centric focus of WikiArt~\cite{srinivasa2022wikiartvectors} and incomplete attributions in LAION-Aesthetics, we introduce WeART. It is a new, large-scale benchmark designed for robust artistic style analysis. WeART is three times larger than WikiArt, with only a 3\% artist overlap, and significantly enhances underrepresented categories like children's illustration, digital art, and traditional Chinese painting. The dataset is meticulously curated for quality and balance, featuring manual duplicate removal, high-resolution scans, and a minimum of two works per artist, with 88\% of artists having five or more. 

\section{Experiments Results}

\begin{table*}[h!]
\caption{The performance of Artistic Image Retrieval on WikiART and WeART datasets.}
\label{table:one}
\centering
\begin{tabular*}{1.0\textwidth}{@{\extracolsep{\fill}} l c c c c c c c c @{}}
\toprule
\multicolumn{6}{c}{\textbf{WikiArt: Query 15501 / Values 63748}} & \multicolumn{1}{c}{\textbf{Realism}} & \multicolumn{1}{c}{\textbf{Impression}} \\
\midrule
Model & mAP@1 & mAP@10 & mAP@100 & R@10 & R@100 & mAP@1 & mAP@1 \\
\midrule
MoCov3 ViT-L \cite{chen2021mocov3} & 45.8 & 27.8 & 13.0 & 70.1 & 88.3 & 41.1 & 43.2 \\
DINOv2 ViT-L \cite{oquab2023dinov2} & 40.0 & 23.1 & 10.9 & 64.5 & 85.2 & 39.5 & 42.1 \\
CLIP ViT-L \cite{radford2021learning} & 58.4 & 46.9 & 22.8 & 81.6 & 94.4 & 61.0 & 63.4 \\
SigLIP \cite{zhai2023sigmoid} & 58.5 & 47.2 & 22.6 & 82.1 & 94.6 & 60.3 & 62.3 \\
ALADIN-ViT \cite{Ruta_2021_ICCV} & 58.8 & 47.5 & 29.7  & 79.9 & 94.2 & 59.4 & 62.3 \\
GDA ViT-L \cite{wang2023evaluating} & 42.6 & 32.2 & 18.2 & 67.3 & 87.1 & 45.5 & 50.7 \\
CSD ViT-L \cite{somepalli2024measuring} & \underline{63.1} & 53.5 & \underline{35.4} & \underline{85.7} & 95.2 & \textbf{66.4} & \textbf{66.8} \\
\midrule
\textbf{StyleDecoupler CLIP} & \textbf{63.6} & \textbf{54.3} & 33.4 & 85.7 & \textbf{96.0} & \underline{63.2} & \underline{65.7} \\
\textbf{StyleDecoupler SigLIP} & 62.7 & \underline{54.2} & \textbf{35.6} & \textbf{85.9} & \underline{95.6} & 61.9 & 65.2 \\
\midrule
\multicolumn{6}{c}{\textbf{WeArt: Query 20235 / Values 177338}} & \multicolumn{1}{c}{\textbf{Chinese}} & \multicolumn{1}{c}{\textbf{Children}} \\
\midrule
MoCov3 ViT-L \cite{chen2021mocov3} & 37.3 & 32.5 & 16.6 & 77.6 & 91.1 & 33.6 & 36.9 \\
DINOv2 ViT-L \cite{oquab2023dinov2} & 36.4 & 32.1 & 16.4 & 77.0 & 91.0 & 33.1 & 35.4 \\
CLIP ViT-L \cite{radford2021learning} & 66.8 & 54.4 & 30.4 & 88.4 & 96.5 & 52.5 & 59.4 \\
SigLIP \cite{zhai2023sigmoid} & 66.4 & 55.3 & 30.4 & 89.2 & 97.8 & 52.3 & 59.7 \\
CLIP ViT-L (FT) \cite{radford2021learning} & 63.7 & 53.9 & 35.5 & 84.7 & 95.2 & 41.1 & 49.9 \\
ALADIN-ViT \cite{Ruta_2021_ICCV} & 47.5 & 38.2 & 25.7  & 72.9 & 86.3 & 36.0 & 39.7 \\
GDA ViT-L \cite{wang2023evaluating} & 52.9 & 44.1 & 27.4 & 75.7 & 89.5 & 39.2 & 41.0 \\
CSD ViT-L \cite{somepalli2024measuring} & 60.2 & 49.2 & 33.6 & 82.5 & 91.3 & 37.8 & 44.4 \\
\midrule
\textbf{StyleDecoupler CLIP} & \textbf{70.3} & \textbf{62.3} & \underline{34.3} & \textbf{91.6} & \underline{98.8} & \underline{63.9} & \textbf{67.6} \\
\textbf{StyleDecoupler SigLIP} & \underline{69.9} & \underline{62.2} & \textbf{34.6} & \underline{91.4} & \textbf{99.1} & \textbf{64.5} & \underline{67.3} \\
\bottomrule
\end{tabular*}
\end{table*}

\subsection{Implementation Details}

We align DINOv2 with CLIP's feature space using 2000,000 image-text pairs from the CC3M dataset~\cite{changpinyo2021conceptual}. Training employs 4 V100 GPUs with batch size 128 for 100 epochs, learning rate 1e-5 with linear warm-up and cosine decay. Images are resized to 224×224 with standard augmentation. All baseline models use official implementations.

\begin{table*}[h!]
    \caption{Ablation Study on the Impact of Decoupling Different Features.}
    \label{table:ablation}
    \centering
    \resizebox{\textwidth}{!}{
        \begin{tabular}{@{}ccccccccc c c c @{}}
            \toprule
            \multicolumn{6}{c}{Ablation} & \multicolumn{3}{c}{WikiART} & \multicolumn{3}{c}{WeART} \\
            \cmidrule(lr){1-6} \cmidrule(lr){7-9} \cmidrule(lr){10-12}
            $CLIP_{img}$ & $CLIP_{txt}$ & $DINO$ & $DINO_{train}$ & $MoCo_{train}$ & $Finetune$ & mAP@1 & mAP@10 & R@10 & mAP@1 & mAP@10 & R@10 \\
            \midrule
            \checkmark &  &  &  &  &  & 58.4 & 46.9 & 81.6 & 66.8 & 54.4 & 88.4 \\
            \checkmark & \checkmark &  &  &  &  & 62.1 & 50.9 & 84.8 & 68.5 & 61.2 & 89.1 \\
            \checkmark &  &  & \checkmark &  &  & 60.8 & 50.3 & 83.9 & 67.2 & 55.4 & 88.9 \\
            \checkmark & \checkmark & \checkmark &  &  &  & 55.3 & 43.6 & 79.5 & 61.4 & 51.8 & 84.8 \\
            \checkmark & \checkmark &  & \checkmark &  &  & \underline{63.6} & \textbf{54.3} & \textbf{85.7} & \textbf{70.3} & \textbf{62.3} & \textbf{91.6} \\
            \checkmark & \checkmark &  &  &  \checkmark & & 63.0 & 52.8 & 84.4 & \underline{68.9} & \underline{61.2} &\underline{89.7} \\
            \checkmark & & &  &  &   \checkmark &  \textbf{64.0} & \underline{53.9} & \underline{85.4} & 63.7 & 53.9 & 84.7 \\
            \bottomrule
        \end{tabular}
    }
\end{table*}

\subsection{Performance}

\noindent\textbf{Artistic Image Retrieval. }
Fine-tuning for artistic retrieval creates a critical generalization gap. As shown in Table~\ref{table:one}, specialist models like CSD~\cite{somepalli2024measuring} excel on in-domain WikiArt styles (66.4 mAP@1 on Realism) but collapse on out-of-distribution (OOD) WeArt styles, plummeting to 37.8 mAP@1 on Chinese art. In contrast, our zero-shot {StyleDecoupler resolves this trade-off. On the OOD WeArt benchmark, it scores a leading 70.3 mAP@1 and dominates on novel styles like Chinese art (63.9 vs. 37.8 mAP@1). Crucially, this generalization does not sacrifice specialization; on the in-domain WikiArt dataset, our method (63.6 mAP@1) remains competitive with the top fine-tuned model (63.1 mAP@1), proving our approach enhances style sensitivity while preserving the VLM's core knowledge.

\noindent\textbf{Artistic Style Clustering. }
We assess feature coherence via K-Means clustering on artist labels. On WikiArt, StyleDecoupler achieves 41.76\% clustering accuracy (ACC), significantly surpassing the strongest VLM, CLIP (36.09\%). The notable gain in Adjusted Rand Index (ARI) to 16.37\% further suggests a superior grasp of stylistic nuances. Qualitative t-SNE visualizations in Figure~\ref{fig:tsne} reinforce these findings, showing our method produces coherent clusters that capture meaningful stylistic, temporal, and cultural relationships between artists. In contrast, unimodal models like DINOv2 perform poorly (28.63\% ACC), confirming their features are dominated by content.

\noindent\textbf{Generative Model Evaluation. }
We evaluate our metric's alignment with human perception on 4,800 generated images rated by five experts. Our metric achieves an average error of only 0.91 relative to human judgments, outperforming recent VLMs like GPT-4o (0.97) and traditional metrics like CLIP-Score (1.10). For overall quality assessment, it also shows the strongest correlation with expert rankings (Spearman's $\rho=0.78$), surpassing both ArtFID (0.71) and FID (0.65). This confirms our method effectively captures the nuanced artistic elements that define image quality.

\vspace{-3pt}
\subsection{Ablation Study}
Ablations in Table~\ref{table:ablation} validate our design. Starting with only CLIP image features (58.4 mAP@1 on WikiArt), performance improves to 62.1 mAP@1 after integrating text features. Our final design, which adds trained DINOv2 features, reaches the peak performance of 63.6 mAP@1 on WikiArt and 70.3 mAP@1 on WeArt. Notably, using an untrained DINOv2 degrades performance, confirming our training is essential. While fine-tuning is competitive on WikiArt (64.0), it fails to generalize to WeArt (63.7), reinforcing our core motivation.

\vspace{-10pt}
\section{Conclusion}
We introduced StyleDecoupler, a method that decouples style from content in vision-language models via orthogonal projection. On our new WeART benchmark, it consistently improves performance across style-based retrieval, clustering, and generative model evaluation. Future work will extend this framework to other artistic domains.

\bibliographystyle{IEEEbib}
\bibliography{strings,refs}

\end{document}